\pdfoutput=1

\documentclass[11pt]{article}

\usepackage{acl}

\usepackage{times}
\usepackage{latexsym}
\usepackage{enumitem}

\usepackage[T1]{fontenc}

\usepackage[utf8]{inputenc}
\usepackage{amsmath}
\usepackage{microtype}
\usepackage{graphicx}
\usepackage{multirow}
\usepackage{booktabs}
\usepackage{xcolor}
\usepackage{color, colortbl}
\definecolor{Gray}{gray}{0.9}

\usepackage{inconsolata}

\title{Benchmarking Large Language Models on Communicative Medical Coaching: A Dataset and a Novel System}

\author{Hengguan Huang\textsuperscript{1}\thanks{\ \ Equal contribution. Correspondence to: Hengguan Huang <\texttt{huang.hengguan@u.nus.edu>}},~
       Songtao Wang\textsuperscript{1}\footnotemark[1],~
       Hongfu Liu\textsuperscript{1},~
       Hao Wang\textsuperscript{2},~
       Ye Wang\textsuperscript{1}
       \\\\
       \textnormal{\textsuperscript{1}National University of Singapore},  
       \textnormal{\textsuperscript{2}Rutgers University}
      }      

\begin{document}

\maketitle
\begin{abstract}
Traditional applications of natural language processing (NLP) in healthcare have predominantly focused on patient-centered services, enhancing patient interactions and care delivery, such as through medical dialogue systems. However, the potential of NLP to benefit inexperienced doctors, particularly in areas such as communicative medical coaching, remains largely unexplored. We introduce ``ChatCoach,'' a human-AI cooperative framework designed to assist medical learners in practicing their communication skills during patient consultations. ChatCoach \footnote{Our data and code are available online: \url{https://github.com/zerowst/Chatcoach} } differentiates itself from conventional dialogue systems by offering a simulated environment where medical learners can practice dialogues with a patient agent, while a coach agent provides immediate, structured feedback.
This is facilitated by our proposed Generalized Chain-of-Thought (GCoT) approach, which fosters the generation of structured feedback and enhances the utilization of external knowledge sources. Additionally, we have developed a dataset specifically for evaluating Large Language Models (LLMs) within the ChatCoach framework on communicative medical coaching tasks. Our empirical results validate the effectiveness of ChatCoach.


\end{abstract}

\section{Introduction}
The advent of Natural Language Processing (NLP) has significantly impacted the healthcare domain, carving pathways for numerous applications that enhance both patient-centered services and healthcare operations. These applications encompass medical dialogue systems, automated medical coding, clinical decision support, and information extraction from electronic health records, among others \cite{he2023survey}. Despite the strides made in these areas, there remains largely untapped potential of NLP in aiding the professional development of early-stage medical learners and early-career practitioners. A critical aspect of this professional development revolves around enhancing communication skills, especially in the context of medical consultations.


A wealth of research highlights the critical importance of effective communication in medical practice. \citet{choudhary2015teaching} found a strong consensus among medical students on the need to refine communication skills for better medical practice, with a significant proportion showing marked improvements after training. Various other studies \cite{choudhary2015teaching,chi2014icap,ruiz2006impact,sargeant2010processes} have consistently shown that proficient communication skills are key to increasing patient satisfaction, enhancing diagnostic precision, and fostering stronger doctor-patient relationships. Despite this acknowledgment, the area of communicative medical coaching, particularly through leveraging advanced Language Language Models (LLMs), remains relatively unexplored. 


Addressing this gap, we introduce \emph{ChatCoach}, a novel human-AI cooperative framework devised to enhance communicative proficiency among medical learners. Unlike traditional dialogue systems focused on \emph{patient} engagement, ChatCoach transitions the focus towards the professional development of \emph{medical practitioners}.  This approach fosters a dynamic environment where learners can engage in realistic dialogues, receive immediate feedback, and refine their understanding of medical terminologies. ChatCoach provides a simulated, realistic environment for medical learners to practice their communication skills during patient consultations. The architecture of ChatCoach (shown in Fig. \ref{fig:overview}(a)) includes a patient agent simulating real-world doctor-patient interactions and a coach agent providing real-time feedback on learners' terminological usage.

A major challenge in this direction is the absence of publicly available data for communicative medical coaching, largely due to the sensitive nature of healthcare information and the substantial costs of data collection and annotation. To overcome this challenge, we devised a multi-agent data generation framework (shown in Fig. \ref{fig:overview}(b)) using external resources to produce training data for fine-tuning an open-source LLM. This framework employs LLM-based agents, including patient, coach, and doctor agents, which interact by querying and retrieving information from two sources: a medical dialogue database and a medical knowledge database. Additionally, we compiled a human-annotated testing dataset to assess LLMs' capabilities in communicative medical coaching.



Our contributions are threefold:

\begin{itemize}
\item  We pioneer the utilization of LLMs for communicative coaching in healthcare, forging a novel intersection among education, healthcare, and AI.
\item  We introduce the first benchmark dataset and evaluation metrics for communicative medical coaching, enabling the assessment of LLMs coaching efficacy in a simulated practice environment.
\item We present a new prompting strategy, dubbed as Generalized Chain-of-Thought (GCoT), devised to improve the generation of structured feedback and the incorporation of external knowledge, without the need for manually constructing reasoning steps. Our GCoT method demonstrates superior performance over various existing Chain-of-Thought techniques across tasks within our dataset.
\end{itemize}

\section{Related Work}

\subsection{ Medical NLP Applications with LLM  }
The field of healthcare has seen notable changes in recent years, driven in part by advances in Natural Language Processing (NLP) technologies. Initially, research efforts were concentrated on fundamental tasks such as Named Entity Recognition (NER) \cite{zhang2021cblue,nesterov2022distantly}, Relation Extraction (RE) \cite{deng2020multimodal,zhao2022hingrl}, and Electronic Health Records (EHR) \cite{yu2019biobert}. These tasks posed challenges due to limited data access and the intricate nature of the medical domain. However, with the emergence of large language models (LLMs), the focus has shifted towards more practical applications, including the development of medical dialogue systems \cite{dou2023plug,qin2023read}, innovative medical consultation platforms \cite{shi2023midmed}, and automated generation of medical reports \cite{zhao2023chatcad+}.
Despite the strides made, the majority of existing models and tools primarily cater to patient-centered services. Notably absent are resources tailored for inexperienced medical learners and early-career doctors, a gap that our research seeks to address. This work delves into the potential of LLMs in enhancing the communication skills of medical professionals. 

\subsection{ Medical Education with NLP }
Traditional techniques aimed at enhancing communication skills include computer-assisted language learning \cite{levy1997computer}, pronunciation training \cite{li2016mispronunciation}, and mispronunciation localization \cite{wei2022unsupervised}. These approaches typically rely on advanced acoustic models \cite{mohamed2011acoustic, huang2019recurrent, huang2020deep, huang2021strode} to identify pronunciation errors and generate feedback. However, these applications are generally designed for the broader public and may not be ideally suited for clinical environments.

In a different vein of research, studies such as \cite{denny2003understanding, da2010corpus, zhang2012automated, chary2019review} have employed NLP techniques to enhance medical education by focusing on content analysis and student performance evaluation. Unlike these approaches, the current work introduces real-time coaching in communication skills specifically tailored for medical consultations. Utilizing Large Language Models (LLMs), it offers immediate, structured feedback, distinguishing itself from the predominantly static and retrospective analyses found in previous work.


\begin{figure*}[t]
\begin{center}
\vskip -0.5cm
\includegraphics[width=0.85\linewidth]{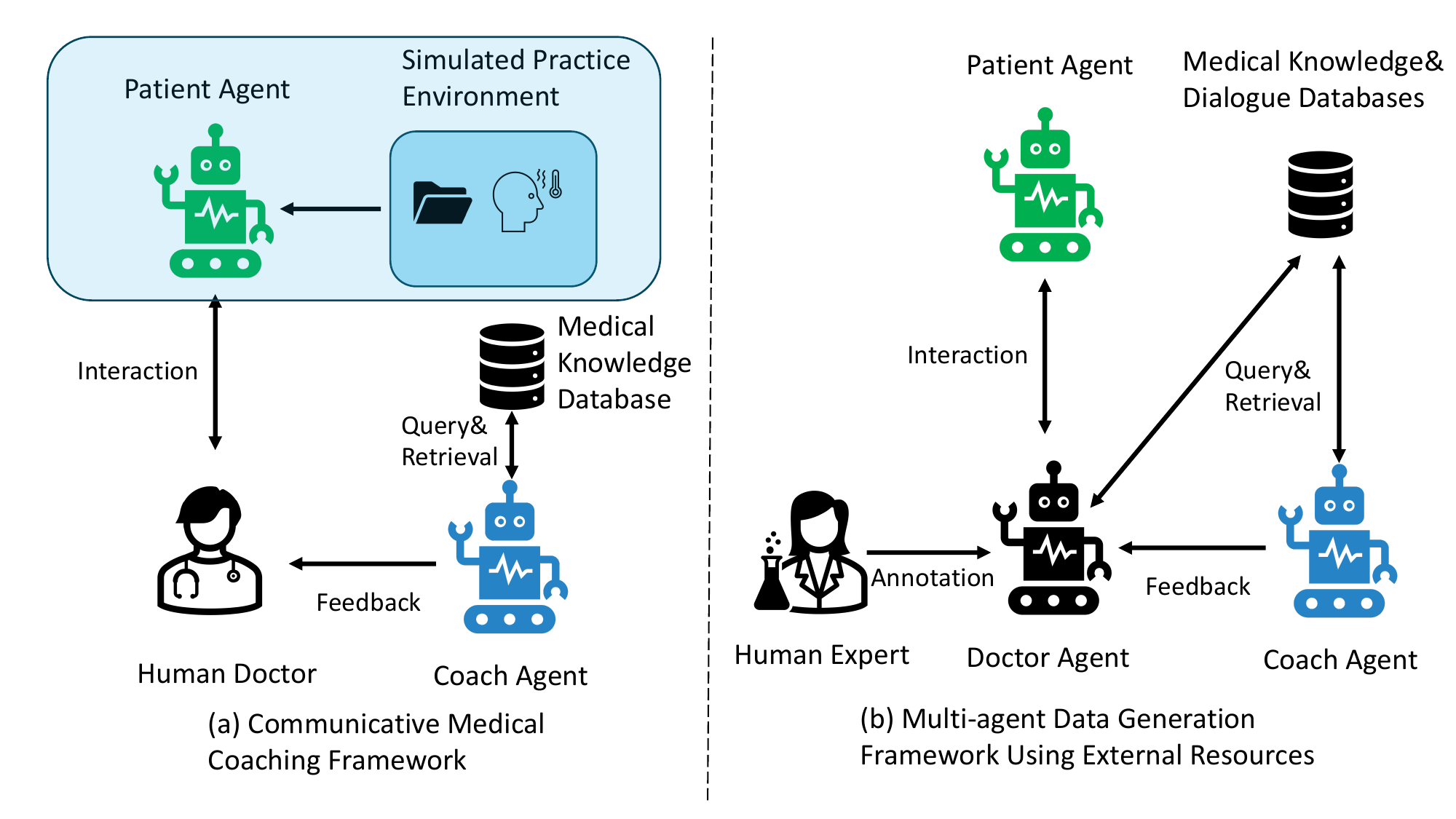} 
\end{center}
\vskip -0.4cm
 \caption{(a) General framework of communicative medical coaching. (b) Multi-agent data generation framework using external resources. }
\label{fig:overview}
\end{figure*}

 \subsection{Prompting-based Method}
Prompting-based methods in Large Language Models (LLMs) have emerged as a versatile mechanism to guide models towards task-specific responses. Among the various strategies, in-context learning \cite{brown2020language}, where relevant examples are provided to tailor the model's behavior, and instruction prompting \cite{wang2022self,ouyang2022training}, where explicit task instructions are embedded within the prompts, have gained prominence. A notable advancement in this domain is the Chain-of-Thought (CoT) paradigm \cite{wei2022chain}, which introduced a chain of reasoning steps for each exemplar of in-context learning, significantly enhancing performance on complex reasoning tasks.  Despite its advancements, CoT's reliance on human-crafted reasoning paths limits its applicability in open-ended settings, such as ours. 

Following this, a variety of strategies have been proposed to improve upon the CoT paradigm. For instance, the zero-shot CoT \cite{kojima2022large} extends the CoT paradigm to handle tasks by simply adding ``think step by step'' to the prompt, without requiring any exemplars and reasoning steps. 
However, such a method does not adequately integrate external knowledge or produce structured feedback that professionals can easily interpret, as observed in our human evaluation. In contrast, our GCoT introduces generalizable variables into the reasoning paths, enabling the generation of structured feedback and the effective integration of external knowledge. Additionally, the development of Auto-CoT \cite{zhang2022automatic, shum2023automatic} aims to lessen the manual burden associated with formulating reasoning steps. However, this method's reliance on generating multiple samples from LLMs introduces computational inefficiencies and falls short in scenarios necessitating immediate feedback, such as our problem settings. This highlights the pressing need for solutions like GCoT that cater to real-time application requirements while enhancing the integration of external knowledge sources.


\section{Communicative Medical Coaching}

\subsection{Problem Formulation}
Given a medical knowledge database \(\mathcal{D}\), which consists of a set of diseases \(D = \{d_k \mid k = 1, \ldots, K\}\), where each \(d_k\) includes a comprehensive description of the disease involving symptoms, medications, and other relevant clinical information. We define the simulated medical environment as \(\mathcal{E}\), comprising a collection of scenarios \(\mathcal{E} = \{e_j \mid j = 1, \ldots, J\}\). Each scenario \(e_j = \{ 
p_j, D_j \} \) corresponds to a \emph{patient agent}, which encapsulates a patient profile \(p_j\) and a specific medical context drawn from a subset of diseases \(D_j \subseteq D\). The goal is to construct a simulated practice environment where a human doctor (i.e., a medical learner) can engage in medical dialogue with a \emph{patient agent}. Concurrently, a \emph{coach agent} delivers real-time feedback to the doctor. 


\subsection{System Overview}

Figure \ref{fig:overview}(a) shows the architecture of the proposed system. It consists of two primary components:  a patient agent, and a coach agent. The patient agent and the coach agent are driven by LLMs.
The human doctor interacts with the patient agent in a simulated medical environment that is specified by each unique scenario \( e_j \).

The patient agent generates responses $\mathcal{R}_j$ during the consultation based on the patient profile \( p_j \), the current input from the doctor \( S_j \), and the preceding part of the conversation \( \mathcal{H}^{-}_{j} \):
\begin{equation} 
g(S_j, p_j, \mathcal{H}^{-}_{j}) \rightarrow \mathcal{R}_j,
\end{equation} 
where \( \mathcal{H}^{-}_{j} \) represents the historical dialogue excluding interactions from the coach agent, ensuring that the coach agent's contributions do not affect the patient agent's responses.

Simultaneously, the coach agent monitors the dialogue between the human doctor and the patient agent, ready to provide feedback. This feedback mechanism is written as:
\begin{equation} 
f(S_j, e_j, \mathcal{H}_j) \rightarrow \mathcal{F}_j ,
\end{equation} 
where \( f \) processes the doctor's dialogue \( S_j \) and the complete history of the conversation \( \mathcal{H}_j \) (including the coach agent's feedback in the previous dialogue round) within the context of \( e_j \) to generate feedback \( \mathcal{F}_j \). The purpose of the coach agent's feedback is to foster improved communication strategies by the doctor, such as correcting errors in medical terminology and providing constructive guidance and encouragement for more effective patient interactions. Both \( g \) and \( f \) are implemented by prompting LLMs. 

\begin{table*}[htb!] 
    \centering
    \begin{tabular}{p{0.95\linewidth}}
    \toprule
    \rowcolor{Gray} \textbf{Generalizable Variables Inferred by GCoT}\\ \midrule
    \textbf{Generalizable Variables across Examples:}

\textit{Condition Miscommunication:}
\begin{itemize}[nosep]
    \item Incorrect disease name or symptom.
    \item Correct disease name or symptom based on medical context.
\end{itemize}

\textit{Medication Miscommunication:}
\begin{itemize}[nosep]
    \item Incorrect medication name or treatment suggestion.
    \item Correct medication name or treatment suggestion based on medical context.
\end{itemize}

\textit{Treatment Miscommunication:}
\begin{itemize}[nosep]
    \item Incorrect treatment advice.
    \item Correct treatment advice based on medical context.
\end{itemize}
%
\\
\bottomrule
    \end{tabular}
    \caption{Generalizable variables inferred by GCoT.}
    \label{tab:gcot_v}
\end{table*} 

\section{Generalized Chain-of-Thought (GCoT)}

Communicative medical coaching poses a unique challenge for Large Language Models (LLMs), characterized by its open-ended, knowledge-intensive reasoning demands.  The feedback generated must be real-time and easily understandable by medical practitioners. Additionally, the reasoning process requires the utilization of external knowledge databases. Traditional prompting methods, such as zero-shot CoT often fall short in generating structured feedback and effectively incorporating external knowledge (refer to Fig. \ref{fig:case_study} for an example of coach feedback generated by prompting-based approaches). Here, we introduce the Generalized Chain-of-Thought (GCoT) approach. GCoT improves upon CoT by embedding generalizable variables within reasoning paths. These variables are elements shared across various data samples' reasoning steps,  facilitating the creation of structured feedback and seamless external knowledge integration.

GCoT adopts a two-step process aimed at utilizing generalizable variables for prompt generation:
\begin{enumerate}
    \item \textbf{Inferring Generalizable Variables across Data Samples:} The process begins with extracting generalizable variables from various input-output samples. This is accomplished by prompting an LLM with: ``Imagine you are reasoning step by step from input to output, please infer generalizable variables in the reasoning steps across the following data samples.'' The input includes the doctor's statement and medical context from a medical knowledge database, with the output being the coach's feedback. This step is critical for identifying variables that represent both the conversation structure and the external knowledge sources, as depicted in Table \ref{tab:gcot_v}.
     \item \textbf{Prompt Generation Based on Inferred Variables:} After identifying these variables, the next step involves generating tailored prompts. The LLM is instructed with: ``Generate the corresponding prompt for GPT-3.5, which should: (1) follow the Chain-of-Thought patterns; (2) ensure reasoning steps are not specific to any data; (3) base reasoning steps on these variables.'' 
     This ensures the feedback (1) adheres to CoT step-by-step reasoning pattern, (2) abstracts reasoning steps for various data samples, and (3) incorporates the identified variables. The outcomes of this prompt generation are documented in Table \ref{tab:gcot}, with variables indicated in square brackets, for instance, [incorrect symptom]. 
\end{enumerate}

\begin{table*}[htb!]
    \centering
    \begin{tabular}{p{0.95\linewidth}}
    \toprule
    \rowcolor{Gray} \textbf{Generalized Chain-of-thought (GCoT)}\\ \midrule
    \textbf{Instruction}:
     As a linguistic coach for a junior doctor, your task is to evaluate the doctor's statement: \{doctor’s statement\} against the provided medical context: \{Medical Context\}. Your evaluation should identify any discrepancies within the doctor's communication. Where discrepancies arise, guide the doctor towards more accurate medical terminology and understanding. If the statements align well with the medical context, provide positive reinforcement and additional advice if necessary. \\
    \textbf{Thinking steps:} \\
    \textit{Identify Key Medical Terms:} \\ Extract medical terms from the doctor's statement, including diseases, symptoms, medications, and treatments.\\
\textit{Compare with Medical Context:}  \\Check these terms against the medical context for accuracy in:
\begin{itemize}[nosep]
    \item Disease/symptom identification.
    \item Medication/treatment recommendation.
\end{itemize}
\textit{Feedback:} 
\begin{itemize}[nosep]
    \item \textit{If Incorrect:} Point out the error and provide the correct term from the medical context. Use simple corrections like ``Instead of [incorrect symptom], it should be [correct symptom]'', ``Instead of [incorrect medication name], it should be [correct medication name]'' or ``Instead of [incorrect disease name], it should be [correct disease name]''.
    \item \textit{If Correct:} Affirm with "Your diagnosis/treatment aligns well with the medical context. Good job."
\end{itemize}
Note: <correct symptom>, <correct medication name> and <correct disease name> are extracted from medical context \\
   
    \bottomrule
    \end{tabular}
    \caption{GCoT prompt for ChatCoach. }
    \label{tab:gcot}
\end{table*}

\section{Constructing the ChatCoach Dataset: A Multi-Agent Approach for Generating Domain-Specific Conversational Data} \label{data}

The development of novel NLP applications, especially in specialized fields such as medical coaching, is hindered by the scarcity of domain-specific conversational datasets.   To bridge this gap, we introduce a novel multi-agent data generation framework (illustrated in Fig. \ref{fig:overview}(b)) leveraging external resources to synthetically produce training data for fine-tuning open-source LLMs. 

Our framework integrates three types of LLM-based agents -- Patient, Coach, and Doctor -- each designed to simulate real-world roles within medical dialogues. These agents interact through querying and retrieving information from two primary sources: a medical dialogue database and a medical knowledge database. Specifically, the Patient Agent simulates patient inquiries; the Doctor Agent generates medical responses, potentially including common errors; and the Coach Agent offers corrective feedback or encouragement, drawing from the medical knowledge database. 

More importantly, to rigorlously evaluate the LLM's performance in medical coaching, we compiled a human-annotated testing dataset based on the aforementioned data.

\paragraph{Data Generation Conditioned on External Resources}
Recent studies, such as \citet{jentzsch2023chatgpt}, have identified limitations in current LLMs' ability to generate diverse and contextually rich data samples. Our methodology addresses these limitations by conditioning the data generation process on external resources: the Disease database\cite{yunxiang2023chatdoctor}, which encompasses comprehensive disease-related information (e.g., symptoms, diagnostic tests, treatments, and medications), and the MedDialog database\cite{he2020meddialog}, a corpus of real-world medical consultations. The inclusion of a coaching role, absent in MedDialog, and the simulation of doctor's errors—uncommon in existing dialogues—pose unique challenges, which our framework overcomes by initiating data generation with patient queries from the MedDialog dataset. The Doctor Agent intentionally incorporates common misconceptions to simulate early-stage medical training errors. The Coach Agent then evaluates these responses against accurate medical statements, correcting terminological inaccuracies and enriching the dialogue with supportive insights for diagnostic reasoning.

\subsection{Task Descriptions}
The ChatCoach Dataset aims to benchmark LLMs' medical coaching efficacy, facilitating the development of communicative medical coaching tools for early-career doctors. We introduce two key tasks for assessing the quality of generated coaching feedback:

\begin{itemize}
    \item \textbf{Detection of  Medical Terminology Misuse:} This task involves identifying incorrect medical terminology in the doctor's responses, such as inappropriate disease diagnoses, irrelevant symptoms, or incorrect medication or test usage. Success depends on analyzing conversational history and applying relevant medical knowledge.
    \item \textbf{Correction of Medical Terminology Misuse:} Following the Detection Task, this task focuses on providing corrective advice to address any identified terminology misuse. It similarly requires a deep understanding of conversational context and medical knowledge.
\end{itemize}

Evaluating the coach's feedback for constructiveness, knowledgeability, and clarity is also crucial, although these aspects present quantification and evaluation challenges. We plan to explore these dimensions in future work.

\paragraph{Human Annotation}
For initial annotation, we engage 2-3 annotators (either medical professionals or knowledgeable students) to review doctor responses within 500 conversations, including patient, doctor, and coach interactions. Annotations focus on the detection and correction of medical terminology misuse, with coach feedback serving as a reference. To ensure quality, we manually validate each annotation, utilizing advanced LLMs like GPT-4 to calculate inter-rater agreement rates. We pay special attention to conversations with low agreement, assessing the plausibility, relevance, and completeness of annotations. Ultimately, from the initial 500 conversations, we retain 291 based on rigorous evaluation criteria.
\begin{table}[h]
\centering
\begin{tabular}{lc}
\hline
\textbf{Statistics} & \textbf{Number} \\
\hline
Total conversation & 291 \\
Disease & 99 \\
Doctor's statement & 1,315 \\
Patient's response & 1,315 \\
Condition & 166 \\
Disease & 98 \\
Medication & 39 \\
Treatment case & 295 \\
Correction case & 291 \\
Nonlingual case & 1,015\\
\hline
\end{tabular}
\caption{Statistics of Testing set.}
\label{tab:human_testing_set}
\end{table}
\subsection{Dataset Overview}
The ChatCoach dataset comprises 3,500 conversations with 13,666 utterances, based on real-world medical consultations from the MedDialog dataset. The dataset is divided into training (2,509 conversations), validation (700), and testing (291) sets. While training data may not be essential for benchmarking closed-source LLMs like ChatGPT, it is crucial for fine-tuning less capable models. The testing set  covering 99 diseases, thoroughly annotated, serves as the primary resource for evaluating LLMs' medical coaching performance. Detailed statistics of this set are provided in Table~\ref{tab:human_testing_set}, highlighting the distribution of conversations, doctor and patient statements, the numbers of condition, medication, and treatment miscommunication errors, and the categorization of cases into detection, correction, and non-linguistic advice.  Notice that ``Nonlingual cases'' in Table~\ref{tab:human_testing_set}. corresponds to non-linguistic advice expected from the Coach Agent when no direct medical terminology errors occur in the doctor's statement. In such scenarios, the coach might provide encouragement, further medical insights, or advice to progress the diagnostic procedure or maintain the dialogue's flow. 





\begin{table*}[h]
\centering
\begin{tabular}{lcccccc}
\toprule
\multirow{2}{*}{\textbf{Method}} & \multicolumn{3}{c}{\textbf{Detection}} & \multicolumn{3}{c}{\textbf{Correction}} \\
 & \textbf{BLEU-2} & \textbf{Rouge-L} & \textbf{BERTScore} & \textbf{BLEU-2} & \textbf{Rouge-L} & \textbf{BERTScore} \\
\midrule
\multicolumn{1}{c}{Training-based} \\
\midrule
Instruction-Tuning & \textbf{39.8} & 3.0 & \textbf{77.8} & \textbf{4.0} & 1.7 & 59.7\\
\midrule
\multicolumn{1}{c}{Prompting-based} \\
\midrule
Instruction Prompting & 27.4 & {3.3} & 67.6 & 1.4 & 2.1 & {61.6}\\
Vanilla CoT & 17.7 & 2.7 & 64.1 & 0.1 & \textbf{2.3} & 58.1 \\
Zero-shot CoT & 27.6 & 1.9 & 69.0 & 3.0 & 0.9 & 58.8 \\
GCoT (Ours) & \textbf{34.2} & \textbf{3.7} & \textbf{72.4} & 1.6 & {2.0} & \textbf{65.4}\\
\midrule
Human & 76.6 & 6.0 & 90.5 & 33.5 & 3.6 & 84.1\\
\bottomrule
\end{tabular}
\caption{Performance comparison of various methods on the detection and correction of medical terminology errors. }
\label{tab:dialogue_system_metrics}
\end{table*}

\section{Experiments}

We assess the medical coaching capabilities of both an open-sourced LLM and a close-sourced LLM using the proposed ChatCoach frameworks on two tasks specified in Sec. \ref{data}, namely detection and correction of misuse of the medical terminology.
The generated coach feedbacks are evaluated based on both automatic and human evaluation metrics.
\subsection{Experiment Setup}
\paragraph{Baselines}
We investigate the following methods for addressing our problem settings:
\begin{itemize}
    \item Vanilla Instruction Prompting: A method where the LLM is prompted with direct instructions for dialogue generation without further context.
    \item Zero-shot Chain of Thought (CoT) \cite{kojima2022large}: A simple CoT approach where the LLM is prompted with instructions for dialogue generation, being asked to generate a reasoning chain step by step.
    \item Vanilla Chain of Thought \cite{wei2022chain}: An extension of CoT where the model is given a few examples involving the corresponding reasoning path.
    \item Instruction Tuning \cite{longpre2023flan}: A training-based method that includes instructions to the training input-output pairs for fine-tuning LLMs.
\end{itemize}

\paragraph{Evaluation Metrics}
For the quantitative study, since both detection and correction tasks belong to natural language generation, we employ conventional metrics, including BLEU-2, ROUGE-L, and BERTScore. BLEU-2 measures the precision of bi-gram overlaps, offering insights into the lexical accuracy of the generated text against reference answers. ROUGE-L assesses sentence-level similarity, focusing on the longest common subsequence to evaluate structural coherence. BERTScore is used for a semantic similarity assessment, utilizing BERT embeddings to compare the generated outputs and reference texts on a deeper semantic level.

The generated feedback from Coach Agents comprises open-ended natural language text. We adopt GPT-4 to extract the medical terminology errors and the corresponding corrections from the Coach Agents' feedback, then calculate the automated metrics based on the extracted information against human annotations. To further validate whether the automatic metrics-based
on our annotated reference answers align with the
the actual quality of model predictions, we conduct
additional human evaluation.

\paragraph{Implementation Details}
We adopt `gpt-3.5-turbo' for all our prompting-based methods. The prompts for all experiments are detailed in the Appendix. For instruction-tuning, we adopt QLORA \cite{longpre2023flan} to fine-tune a variant of Llama2, named Chinese\_Alpaca2\_LORA\_13B, using 4*A40 GPUs for approximately 9 hours, with a batch size of 64, a learning rate of $2 \times 10^{-4}$, and a maximum of 1000 training steps. ( See Appendix for prompts of our baseline approaches.)

\subsection{Results}

We present the performance of various methods in Table~\ref{tab:dialogue_system_metrics}, focusing on the detection and correction of medical terminology errors. The apparent gap between machine-generated results and human benchmarks in all evaluated metrics signals the inherent challenges within communicative medical coaching. 

\begin{table*}[ht]
\centering
\begin{tabular}{lcc}
\toprule
\textbf{Error Category} & \textbf{Zero-shot CoT} & \textbf{GCoT} \\
\midrule
Overly Divergent Advice & 7.14 & 0.79 \\
Excessive Coaching & 5.56 & 3.97 \\
Limited Medical Knowledge & 5.56 & 1.59 \\
Role Mismatch & 1.59 & 0.00 \\
\bottomrule
\end{tabular}
\caption{Error rate (\%) comparison between zero-shot CoT and GCoT (ours).}
\label{tab:error}
\end{table*}

\paragraph{Detection of Medical Terminology Misuse}
In terms of the detection task, the results demonstrate GCoT's effectiveness in identifying medical terminology errors, with our method achieving competitive scores in BLEU-2 (34.2), Rouge-L (3.7), and BERTScore (72.4) metrics.  Despite the Instruction-tuning method's higher scores in some metrics, GCoT's performance stands out among other prompting methods, indicating its effectiveness without the need for additional fine-tuning.

\paragraph{Correction of Medical Terminology Misuse}
In the correction task, although Instruction-Tuning continues to lead in performance with a BLEU-2 score of 4.0, the gap narrows, indicating the intrinsic challenge associated with generating contextually accurate corrections. When evaluating with the BERTScore, GCoT showcases its strength with a notable BERTScore of 65.4, surpassing both the prompting-based method and the training-based method. This discrepancy indicates that Instruction-Tuning, despite its effectiveness in generating responses that structurally follow given response patterns, may not fully capture the semantic nuances required for the diverse range of correct responses. This might be due to the method's tendency to overfit to the examples within the training dataset, limiting its ability to generalize to the varied corrections encountered in the test set.

\paragraph{Human Evaluation}

To validate the previously observed results, we conducted a human evaluation. We randomly selected 10\% (126 instances) of Testing set for this purpose. Feedback generated by both Baseline Zero-shot CoT and our GCoT was reviewed by two participants, who were asked to rate each piece of feedback on a scale from 1 to 4, with respect to constructiveness, clarity, knowledgeability, and overall quality. Table \ref{tab:humanscore} shows the average scores for each criterion. The results clearly indicate that our proposed approach, GCoT, significantly outperforms the baseline Zero-shot CoT, particularly in terms of clarity and constructiveness. This underscores GCoT's effectiveness in producing structured feedback that is easily understandable for users.

\begin{table}[h]
\centering
\begin{tabular}{lcc}
\toprule
Metric & CoT & GCoT (ours) \\
\midrule
Constructiveness & 2.41 & \textbf{2.68} \\
Clarity & 2.15 & \textbf{3.10} \\
Knowledgeability & 2.35 & \textbf{2.39} \\
Overall & 2.21 & \textbf{2.52} \\
\bottomrule
\end{tabular}
\caption{Human evaluation of coach feedback generated by GCoT and Zero-shot CoT.}
\label{tab:humanscore}
\end{table}

\paragraph{Error Analysis}
To delve deeper into the sources of errors within Zero-shot CoT and GCoT implementations, we annotated all instances involved in the human evaluation, categorizing them into four distinct classes:
\begin{itemize}
    \item \textbf{Overly Divergent Advice:} Feedback that is too wide-ranging, long, or off-topic, reducing its effectiveness.
    \item \textbf{Excessive Coaching:} Feedback inappropriately critiques suitable responses for lacking professional jargon.
    \item \textbf{Limited Medical Knowledge:} Errors due to insufficient use or understanding of the medical knowledge database.
    \item \textbf{Role Mismatch:} Instances where feedback shifts from a coach's to a doctor's perspective, misaligning with the intended advisory role.
\end{itemize}

As demonstrated in Table~\ref{tab:error}, the comparison between Zero-shot CoT and GCoT reveals significant improvements across all error categories with our GCoT approach. Notably,  GCoT dramatically reduces the incidence of overly divergent advice from 7.14\% to a mere 0.79\%. Similarly, errors categorized under Limited Medical Knowledge dropped from 5.56\% to 1.59\%.
These results underscore GCoT's capability to generate more targeted and organized feedback while effectively utilizing external medical knowledge.
\begin{figure*}[t]
\begin{center}
\vskip -0.5cm
\includegraphics[width=0.9\linewidth]{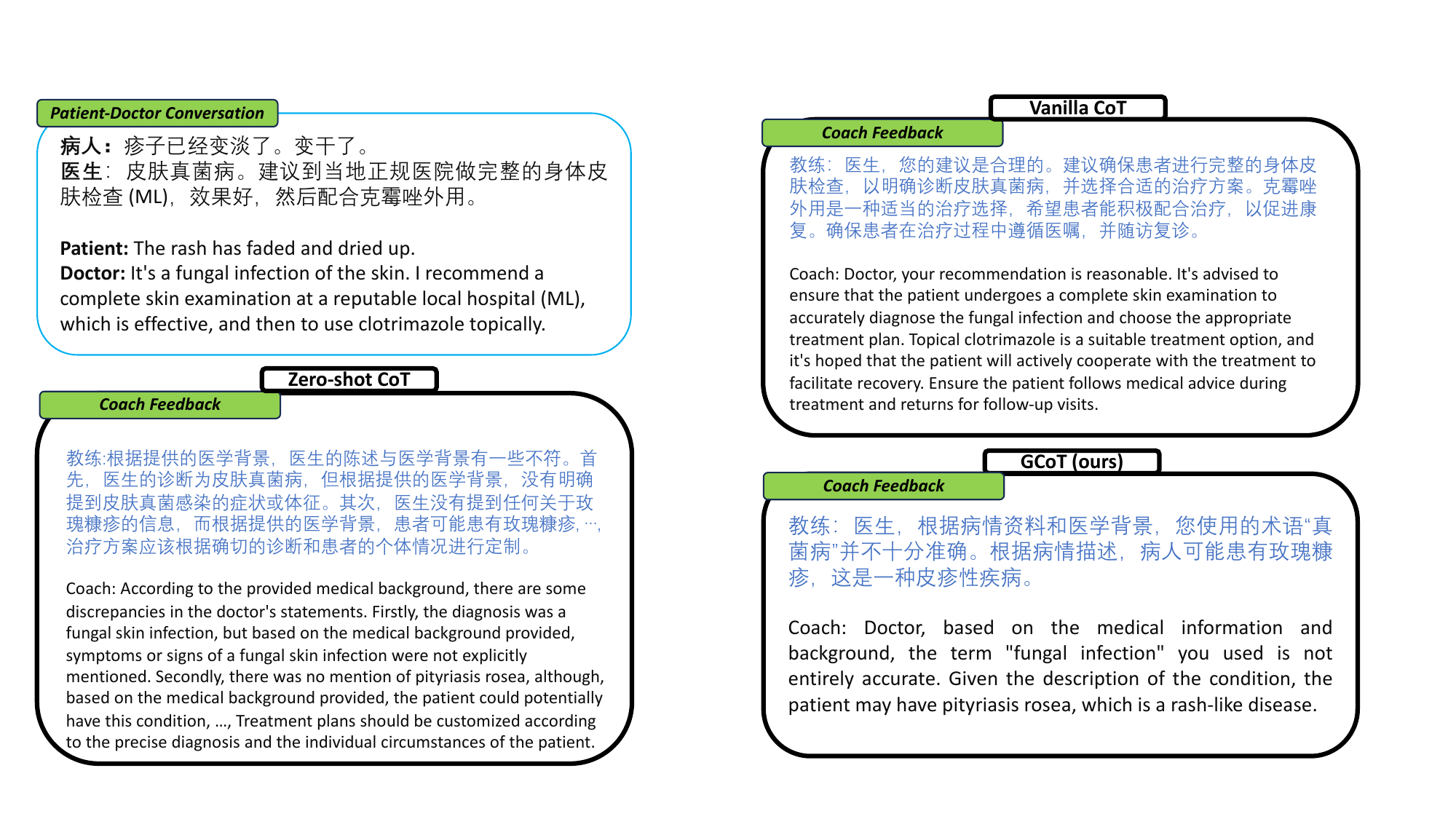} 
\end{center}
\vskip -0.4cm
 \caption{Example of coach feedback generated by various approaches. Vanilla CoT fails to identify errors in medical terminology, possibly due to lacking integration with external knowledge. While thorough, Zero-shot CoT generates overly verbose feedback unsuited for real-time application. In contrast, GCoT identifies errors effectively and provides concise and well-structured feedback, demonstrating superior integration of external medical knowledge for practical real-time coaching. }
\label{fig:case_study}
\end{figure*}

\paragraph{Case Study}

Figure \ref{fig:case_study} showcases an example of coach feedback generated by both baseline prompting methods and our Generalized Chain-of-Thought (GCoT) approach for comparison. In this case, we observe that Vanilla CoT is unable to detect errors in medical terminology, possibly due to its inadequate utilization of external medical knowledge. Zero-shot CoT, on the other hand, produces feedback that is lengthy and circuitous, making it less suitable for the immediacy required in real-time coaching environments. In stark contrast, the example illustrates how GCoT provides feedback that is notably more organized and precise, demonstrating its enhanced ability to integrate external medical knowledge sources effectively.

\section{Conclusion}
This work introduces ChatCoach, a new human-AI cooperative framework for communicative medical coaching. At the core of our approach is the Generalized Chain-of-Thought (GCoT), a strategy that significantly improves feedback structuring and the integration of external knowledge. We developed the first benchmark dataset designed to evaluate the medical coaching capabilities of Large Language Models (LLMs) within the ChatCoach framework. Through a series of automatic and human evaluations, we demonstrate ChatCoach's effectiveness in tackling two key tasks in communicative medical coaching, showcasing its potential to enhance medical education through AI. 

\section*{Limitations}

Despite the advancements made by ChatCoach and the Generalized Chain-of-Thought (GCoT) approach, limitations persist that require further exploration and enhancement. Specifically, our error analysis reveals that GCoT still faces challenges with excessive coaching, where the system may critique acceptable responses for not using professional jargon. This indicates a need for refinement in distinguishing between instances that genuinely require correction. Addressing this issue involves developing an additional component within GCoT that accurately identifies when coach intervention is necessary, thereby reducing unwarranted critiques and enhancing the relevance and precision of the feedback provided.

\section*{Acknowledgments}
The authors would like to thank the anonymous reviewers for their insightful comments and suggestions. This project is funded by a research grant MOE-MOESOL2021-0005
from the Ministry of Education in Singapore. 

\bibliography{custom}

\begin{thebibliography}{36}
\expandafter\ifx\csname natexlab\endcsname\relax\def\natexlab#1{#1}\fi

\bibitem[{Brown et~al.(2020)Brown, Mann, Ryder, Subbiah, Kaplan, Dhariwal, Neelakantan, Shyam, Sastry, Askell et~al.}]{brown2020language}
Tom Brown, Benjamin Mann, Nick Ryder, Melanie Subbiah, Jared~D Kaplan, Prafulla Dhariwal, Arvind Neelakantan, Pranav Shyam, Girish Sastry, Amanda Askell, et~al. 2020.
\newblock Language models are few-shot learners.
\newblock \emph{Advances in neural information processing systems}, 33:1877--1901.

\bibitem[{Chary et~al.(2019)Chary, Parikh, Manini, Boyer, and Radeos}]{chary2019review}
Michael Chary, Saumil Parikh, Alex~F Manini, Edward~W Boyer, and Michael Radeos. 2019.
\newblock A review of natural language processing in medical education.
\newblock \emph{Western Journal of Emergency Medicine}, 20(1):78.

\bibitem[{Chi and Wylie(2014)}]{chi2014icap}
Michelene~TH Chi and Ruth Wylie. 2014.
\newblock The icap framework: Linking cognitive engagement to active learning outcomes.
\newblock \emph{Educational psychologist}, 49(4):219--243.

\bibitem[{Choudhary and Gupta(2015)}]{choudhary2015teaching}
Anjali Choudhary and Vineeta Gupta. 2015.
\newblock Teaching communications skills to medical students: Introducing the fine art of medical practice.
\newblock \emph{International Journal of Applied and Basic Medical Research}, 5(Suppl 1):S41.

\bibitem[{Da~Silva and Dennick(2010)}]{da2010corpus}
Ana~L Da~Silva and Reg Dennick. 2010.
\newblock Corpus analysis of problem-based learning transcripts: an exploratory study.
\newblock \emph{Medical education}, 44(3):280--288.

\bibitem[{Deng et~al.(2020)Deng, Xu, Qiu, Xia, Zhang, and Liu}]{deng2020multimodal}
Yifan Deng, Xinran Xu, Yang Qiu, Jingbo Xia, Wen Zhang, and Shichao Liu. 2020.
\newblock A multimodal deep learning framework for predicting drug--drug interaction events.
\newblock \emph{Bioinformatics}, 36(15):4316--4322.

\bibitem[{Denny et~al.(2003)Denny, Smithers, Miller, and Spickard~III}]{denny2003understanding}
Joshua~C Denny, Jeffrey~D Smithers, Randolph~A Miller, and Anderson Spickard~III. 2003.
\newblock “understanding” medical school curriculum content using knowledgemap.
\newblock \emph{Journal of the American Medical Informatics Association}, 10(4):351--362.

\bibitem[{Dou et~al.(2023)Dou, Jin, Jiao, Zhao, Tao, and Zhao}]{dou2023plug}
Chengfeng Dou, Zhi Jin, Wenping Jiao, Haiyan Zhao, Zhenwei Tao, and Yongqiang Zhao. 2023.
\newblock Plug-and-play medical dialogue system.
\newblock \emph{arXiv preprint arXiv:2305.11508}.

\bibitem[{He et~al.(2023)He, Mao, Lin, Ruan, Lan, Feng, and Cambria}]{he2023survey}
Kai He, Rui Mao, Qika Lin, Yucheng Ruan, Xiang Lan, Mengling Feng, and Erik Cambria. 2023.
\newblock A survey of large language models for healthcare: from data, technology, and applications to accountability and ethics.
\newblock \emph{arXiv preprint arXiv:2310.05694}.

\bibitem[{He et~al.(2020)He, Chen, Ju, Dong, Fang, Wang, Yang, Zeng, Zhang, Zhang et~al.}]{he2020meddialog}
Xuehai He, Shu Chen, Zeqian Ju, Xiangyu Dong, Hongchao Fang, Sicheng Wang, Yue Yang, Jiaqi Zeng, Ruisi Zhang, Ruoyu Zhang, et~al. 2020.
\newblock Meddialog: Two large-scale medical dialogue datasets.
\newblock \emph{arXiv preprint arXiv:2004.03329}.

\bibitem[{Huang et~al.(2021)Huang, Liu, Wang, Xiao, and Wang}]{huang2021strode}
Hengguan Huang, Hongfu Liu, Hao Wang, Chang Xiao, and Ye~Wang. 2021.
\newblock Strode: Stochastic boundary ordinary differential equation.
\newblock In \emph{International Conference on Machine Learning}, pages 4435--4445. PMLR.

\bibitem[{Huang et~al.(2019)Huang, Wang, and Mak}]{huang2019recurrent}
Hengguan Huang, Hao Wang, and Brian Mak. 2019.
\newblock Recurrent poisson process unit for speech recognition.
\newblock In \emph{Proceedings of the AAAI Conference on Artificial Intelligence}, volume~33, pages 6538--6545.

\bibitem[{Huang et~al.(2020)Huang, Xue, Wang, and Wang}]{huang2020deep}
Hengguan Huang, Fuzhao Xue, Hao Wang, and Ye~Wang. 2020.
\newblock Deep graph random process for relational-thinking-based speech recognition.
\newblock In \emph{International Conference on Machine Learning}, pages 4531--4541. PMLR.

\bibitem[{Jentzsch and Kersting(2023)}]{jentzsch2023chatgpt}
Sophie Jentzsch and Kristian Kersting. 2023.
\newblock Chatgpt is fun, but it is not funny! humor is still challenging large language models.
\newblock \emph{arXiv preprint arXiv:2306.04563}.

\bibitem[{Kojima et~al.(2022)Kojima, Gu, Reid, Matsuo, and Iwasawa}]{kojima2022large}
Takeshi Kojima, Shixiang~Shane Gu, Machel Reid, Yutaka Matsuo, and Yusuke Iwasawa. 2022.
\newblock Large language models are zero-shot reasoners.
\newblock \emph{Advances in neural information processing systems}, 35:22199--22213.

\bibitem[{Levy(1997)}]{levy1997computer}
Michael Levy. 1997.
\newblock \emph{Computer-assisted language learning: Context and conceptualization}.
\newblock Oxford University Press.

\bibitem[{Li et~al.(2016)Li, Qian, and Meng}]{li2016mispronunciation}
Kun Li, Xiaojun Qian, and Helen Meng. 2016.
\newblock Mispronunciation detection and diagnosis in l2 english speech using multidistribution deep neural networks.
\newblock \emph{IEEE/ACM Transactions on Audio, Speech, and Language Processing}, 25(1):193--207.

\bibitem[{Longpre et~al.(2023)Longpre, Hou, Vu, Webson, Chung, Tay, Zhou, Le, Zoph, Wei et~al.}]{longpre2023flan}
Shayne Longpre, Le~Hou, Tu~Vu, Albert Webson, Hyung~Won Chung, Yi~Tay, Denny Zhou, Quoc~V Le, Barret Zoph, Jason Wei, et~al. 2023.
\newblock The flan collection: Designing data and methods for effective instruction tuning.
\newblock \emph{arXiv preprint arXiv:2301.13688}.

\bibitem[{Mohamed et~al.(2011)Mohamed, Dahl, and Hinton}]{mohamed2011acoustic}
Abdel-rahman Mohamed, George~E Dahl, and Geoffrey Hinton. 2011.
\newblock Acoustic modeling using deep belief networks.
\newblock \emph{IEEE transactions on audio, speech, and language processing}, 20(1):14--22.

\bibitem[{Nesterov and Umerenkov(2022)}]{nesterov2022distantly}
Alexander Nesterov and Dmitry Umerenkov. 2022.
\newblock Distantly supervised end-to-end medical entity extraction from electronic health records with human-level quality.
\newblock \emph{arXiv preprint arXiv:2201.10463}.

\bibitem[{Ouyang et~al.(2022)Ouyang, Wu, Jiang, Almeida, Wainwright, Mishkin, Zhang, Agarwal, Slama, Ray et~al.}]{ouyang2022training}
Long Ouyang, Jeffrey Wu, Xu~Jiang, Diogo Almeida, Carroll Wainwright, Pamela Mishkin, Chong Zhang, Sandhini Agarwal, Katarina Slama, Alex Ray, et~al. 2022.
\newblock Training language models to follow instructions with human feedback.
\newblock \emph{Advances in Neural Information Processing Systems}, 35:27730--27744.

\bibitem[{Qin et~al.(2023)Qin, Chen, Wang, Lan, Ren, and Hong}]{qin2023read}
Wei Qin, Zetong Chen, Lei Wang, Yunshi Lan, Weijieying Ren, and Richang Hong. 2023.
\newblock Read, diagnose and chat: Towards explainable and interactive llms-augmented depression detection in social media.
\newblock \emph{arXiv preprint arXiv:2305.05138}.

\bibitem[{Ruiz et~al.(2006)Ruiz, Mintzer, and Leipzig}]{ruiz2006impact}
Jorge~G Ruiz, Michael~J Mintzer, and Rosanne~M Leipzig. 2006.
\newblock The impact of e-learning in medical education.
\newblock \emph{Academic medicine}, 81(3):207--212.

\bibitem[{Sargeant et~al.(2010)Sargeant, Armson, Chesluk, Dornan, Eva, Holmboe, Lockyer, Loney, Mann, and van~der Vleuten}]{sargeant2010processes}
Joan Sargeant, Heather Armson, Ben Chesluk, Timothy Dornan, Kevin Eva, Eric Holmboe, Jocelyn Lockyer, Elaine Loney, Karen Mann, and Cees van~der Vleuten. 2010.
\newblock The processes and dimensions of informed self-assessment: a conceptual model.
\newblock \emph{Academic Medicine}, 85(7):1212--1220.

\bibitem[{Shi et~al.(2023)Shi, Liu, Wang, Leng, Xue, Zhang, and Zhang}]{shi2023midmed}
Xiaoming Shi, Zeming Liu, Chuan Wang, Haitao Leng, Kui Xue, Xiaofan Zhang, and Shaoting Zhang. 2023.
\newblock Midmed: Towards mixed-type dialogues for medical consultation.
\newblock \emph{arXiv preprint arXiv:2306.02923}.

\bibitem[{Shum et~al.(2023)Shum, Diao, and Zhang}]{shum2023automatic}
KaShun Shum, Shizhe Diao, and Tong Zhang. 2023.
\newblock Automatic prompt augmentation and selection with chain-of-thought from labeled data.
\newblock \emph{arXiv preprint arXiv:2302.12822}.

\bibitem[{Wang et~al.(2022)Wang, Kordi, Mishra, Liu, Smith, Khashabi, and Hajishirzi}]{wang2022self}
Yizhong Wang, Yeganeh Kordi, Swaroop Mishra, Alisa Liu, Noah~A Smith, Daniel Khashabi, and Hannaneh Hajishirzi. 2022.
\newblock Self-instruct: Aligning language model with self generated instructions.
\newblock \emph{arXiv preprint arXiv:2212.10560}.

\bibitem[{Wei et~al.(2022{\natexlab{a}})Wei, Wang, Schuurmans, Bosma, Xia, Chi, Le, Zhou et~al.}]{wei2022chain}
Jason Wei, Xuezhi Wang, Dale Schuurmans, Maarten Bosma, Fei Xia, Ed~Chi, Quoc~V Le, Denny Zhou, et~al. 2022{\natexlab{a}}.
\newblock Chain-of-thought prompting elicits reasoning in large language models.
\newblock \emph{Advances in Neural Information Processing Systems}, 35:24824--24837.

\bibitem[{Wei et~al.(2022{\natexlab{b}})Wei, Huang, Gu, Wang, and Wang}]{wei2022unsupervised}
Wei Wei, Hengguan Huang, Xiangming Gu, Hao Wang, and Ye~Wang. 2022{\natexlab{b}}.
\newblock Unsupervised mismatch localization in cross-modal sequential data with application to mispronunciations localization.
\newblock \emph{Transactions on Machine Learning Research}.

\bibitem[{Yu et~al.(2019)Yu, Hu, Lu, Sun, and Yuan}]{yu2019biobert}
Xin Yu, Wenshen Hu, Sha Lu, Xiaoyan Sun, and Zhenming Yuan. 2019.
\newblock Biobert based named entity recognition in electronic medical record.
\newblock In \emph{2019 10th international conference on information technology in medicine and education (ITME)}, pages 49--52. IEEE.

\bibitem[{Yunxiang et~al.(2023)Yunxiang, Zihan, Kai, Ruilong, and You}]{yunxiang2023chatdoctor}
Li~Yunxiang, Li~Zihan, Zhang Kai, Dan Ruilong, and Zhang You. 2023.
\newblock Chatdoctor: A medical chat model fine-tuned on llama model using medical domain knowledge.
\newblock \emph{arXiv preprint arXiv:2303.14070}.

\bibitem[{Zhang et~al.(2021)Zhang, Chen, Bi, Liang, Li, Shang, Yin, Tan, Xu, Huang et~al.}]{zhang2021cblue}
Ningyu Zhang, Mosha Chen, Zhen Bi, Xiaozhuan Liang, Lei Li, Xin Shang, Kangping Yin, Chuanqi Tan, Jian Xu, Fei Huang, et~al. 2021.
\newblock Cblue: A chinese biomedical language understanding evaluation benchmark.
\newblock \emph{arXiv preprint arXiv:2106.08087}.

\bibitem[{Zhang et~al.(2012)Zhang, Pakhomov, Gladding, Aylward, Borman-Shoap, and Melton}]{zhang2012automated}
Rui Zhang, Serguei Pakhomov, Sophia Gladding, Michael Aylward, Emily Borman-Shoap, and Genevieve~B Melton. 2012.
\newblock Automated assessment of medical training evaluation text.
\newblock In \emph{AMIA annual symposium proceedings}, volume 2012, page 1459. American Medical Informatics Association.

\bibitem[{Zhang et~al.(2022)Zhang, Zhang, Li, and Smola}]{zhang2022automatic}
Zhuosheng Zhang, Aston Zhang, Mu~Li, and Alex Smola. 2022.
\newblock Automatic chain of thought prompting in large language models.
\newblock \emph{arXiv preprint arXiv:2210.03493}.

\bibitem[{Zhao et~al.(2022)Zhao, Hu, You, Wang, and Su}]{zhao2022hingrl}
Bo-Wei Zhao, Lun Hu, Zhu-Hong You, Lei Wang, and Xiao-Rui Su. 2022.
\newblock Hingrl: predicting drug--disease associations with graph representation learning on heterogeneous information networks.
\newblock \emph{Briefings in bioinformatics}, 23(1):bbab515.

\bibitem[{Zhao et~al.(2023)Zhao, Wang, Gu, Zhu, Mei, Zhuang, Cui, Wang, and Shen}]{zhao2023chatcad+}
Zihao Zhao, Sheng Wang, Jinchen Gu, Yitao Zhu, Lanzhuju Mei, Zixu Zhuang, Zhiming Cui, Qian Wang, and Dinggang Shen. 2023.
\newblock Chatcad+: Towards a universal and reliable interactive cad using llms.
\newblock \emph{arXiv preprint arXiv:2305.15964}.

\end{thebibliography}

\newpage

\appendix

\section{A Failed Example of Coach Feedback}
\begin{figure*}[h]
\begin{center}
\includegraphics[width=0.99\linewidth]{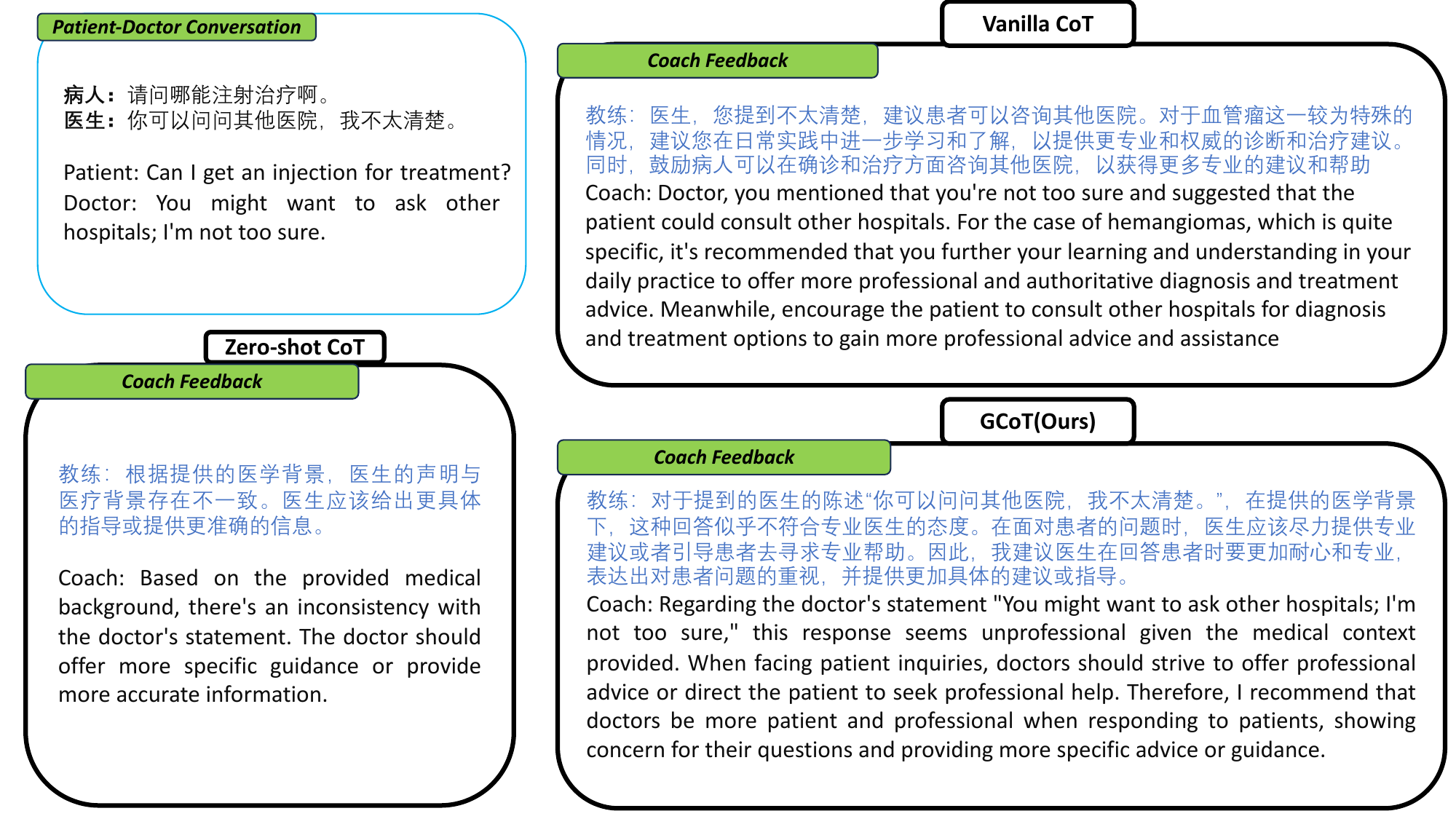} 
\end{center}
 \caption{A failed example of coach feedback from various prompting-based approaches, demonstrating the issue of excessive coaching.}
\label{fig:case_study}
\end{figure*}
\section{Prompts of Our Baseline Approaches}

In this section, we present the prompts used for each baseline approach: Instruction Prompting (see Table \ref{tab:instruction_prompting}), Vanilla Chain-of-Thought (see Table \ref{tab:vanilla_cot}), and Zero-shot Chain-of-Thought (see Table \ref{tab:zero}).

\begin{table*}[htb!]
    \centering
    \begin{tabular}{p{0.95\linewidth}}
    \toprule
    \rowcolor{Gray} \textbf{Vanilla Instruction Prompting}\\ \midrule
    \textbf{Instruction}: As a linguistic coach for a junior doctor, evaluate the doctor's statement: \{doctor’s statement\} against the given medical context: \{medical context\}. If there are discrepancies, guide the doctor. If not, provide positive feedback.  \\
    \bottomrule
    \end{tabular}
    \caption{Instruction prompting for ChatCoach.}
    \label{tab:instruction_prompting}
\end{table*}

\begin{table*}[htb!]
    \centering
    \begin{tabular}{p{0.95\linewidth}}
    \toprule
    \rowcolor{Gray} \textbf{Vanilla Chain-of-thought}\\ \midrule
 \textbf{Instruction}: As a linguistic coach for a junior doctor, evaluate the doctor's statement: \{doctor’s statement\} against the given medical context: \{medical context\}. You should provide your response based on the following examples of input,  thinking steps and output. \\
 \textbf{Example 1}:   \\
 ~~~ \textbf{Input}:   \\
 ~~~~~~ \{doctor’s statement for Example 1\}  \\
  ~~~~~~ \{medical context for Example 1\}  \\
 ~~~ \textbf{Thinking steps}: \\
   ~~~~~~ \{thinking steps for Example 1\}  \\
  ~~~ \textbf{Output}: \\
     ~~~~~~ \{coach's feedback for Example 1\}  \\
 \textbf{Example 2}:  \{example2\} \\
 \textbf{Example 3}:  \{example3\} \\

 ~~~ \textbf{Input}:   \\
 ~~~~~~ \{doctor’s statement\}  \\
  ~~~~~~ \{medical context\}  \\
    \bottomrule
    \end{tabular}
    \caption{Vanilla CoT for ChatCoach.}
    \label{tab:vanilla_cot}
\end{table*}

\begin{table*}[htb!]
    \centering
    \begin{tabular}{p{0.95\linewidth}}
    \toprule
    \rowcolor{Gray} \textbf{Zero-shot Chain-of-thought}\\ \midrule
       \textbf{Instruction}: As a linguistic coach for a junior doctor, evaluate the doctor's statement: \{doctor’s statement\} against the given medical context: \{medical context\}. If there are discrepancies, guide the doctor. If not, provide positive feedback.  \\ \textbf{Please think step by step.} \\
    \bottomrule
    \end{tabular}
    \caption{Zero-shot CoT for ChatCoach}
    \label{tab:zero}
\end{table*}



\end{document}